\documentclass[letterpaper, 10 pt, conference]{ieeeconf}  

\IEEEoverridecommandlockouts                              

\usepackage{graphicx} 
\usepackage{epsfig} 
\usepackage{mathptmx} 
\usepackage{times} 
\usepackage{amsmath} 
\usepackage{amssymb}  
\usepackage{hyperref}

\usepackage{todonotes}

\title{\LARGE \bf The Urban Last Mile Problem: \\ Autonomous Drone Delivery to Your Balcony}

\author{Gino Brunner, Bence Szebedy, Simon Tanner and Roger Wattenhofer*
\thanks{*Computer Engineering and Networks Laboratory, ETH Zurich.}%
}

\begin{document}

\maketitle
\thispagestyle{empty}
\pagestyle{empty}


\begin{abstract}
Drone delivery has been a hot topic in the industry in the past few years. However, existing approaches either focus on rural areas or rely on centralized drop-off locations from where the last mile delivery is performed. In this paper we tackle the problem of autonomous last mile delivery in urban environments using an off-the-shelf drone. We build a prototype system that is able to fly to the approximate delivery location using GPS and then find the exact drop-off location using visual navigation. The drop-off location could, e.g., be on a balcony or porch, and simply needs to be indicated by a visual marker on the wall or window. We test our system components in simulated environments, including the visual navigation and collision avoidance. Finally, we deploy our drone in a real-world environment and show how it can find the drop-off point on a balcony. To stimulate future research in this topic we open source our code.

\end{abstract}


\section{Introduction} \label{sec:Intro}

Amazon, Google, UPS 
and a multitude of startups are testing drone delivery services. They all have impressive videos that show their systems in action. These videos have in common that the drones are delivering a package to a farm or an estate, with no other house in sight. In reality, a growing number of people in the world live in dense urban areas, in apartments or condominiums. According to World Bank data, urban living is the norm already today, and its share is growing steadily~\cite{worldbankurbanpop}.

In this paper we study the \emph{urban} last mile problem. In a multi-storey house environment, the problem does not end with finding the right building. As a next step the drone needs to find the right apartment. Then, the drone must drop the packet at the right location, usually a balcony with limited space for maneuvering. While doing all this, the drone needs to circumnavigate any obstructions. 

Our contribution is a suite of software\footnote{Available here: \\ \url{https://github.com/szebedy/autonomous-drone}} aiming to enable autonomous last mile drone delivery, including: 

\begin{itemize}

\item Autonomous control logic with high level control functions (takeoff, land, approach visual marker, fly to coordinates, etc.) and interfacing the rest of the software nodes.

\item Visual odometry algorithm (SVO) with full calibration for the Intel Aero RTF Drone, usable for vision based navigation.

\item Trajectory planner node (Ewok) with collision avoidance, adjusted for usage with the Intel Aero RTF Drone.

\item Visual marker tracker (WhyCon) providing precise estimates of the marker location at a distance of up to 7 meters using the front-facing camera of the Intel Aero RTF Drone with a resolution 640 by 480 pixels.

\item Simulation environment with a preconfigured test scenario including an apartment with balcony equipped with the WhyCon visual marker. The environment also contains a simulated drone that seamlessly replaces real hardware by providing the same output messages.

\end{itemize}

\section{Related work}

Despite the simultaneously growing popularity of unmanned aerial vehicles (UAVs) and online shopping with home delivery, the amount of scientific work in the area of drone delivery is surprisingly limited. Although Joerss et al.~\cite{Joerss} predict that autonomous vehicles will deliver 80 percent of parcels in the future, they conclude that an autonomous drone delivery model is only viable in rural areas. D'Andrea~\cite{Andrea} states that although drone delivery can be feasible and even profitable, there are a few key challenges that need to be addressed, like vehicle design, localization, navigation and coordination. Additional challenges include privacy, security and government regulations.

Guerrero et al.~\cite{Guerrero} propose a UAV design with a cable suspended payload. They also design control laws such that the swing of the payload is minimized along the trajectory. In order to address vehicle coordination, Dorling et al.~\cite{Dorling} propose two vehicle routing problems, one of which minimizes delivery costs, while the other minimizes the overall delivery time. A complete prototype system is proposed in \cite{Gatteschi}, including an Android application to place orders of drug parcels and an autonomous drone to deliver them. Park et al.~\cite{Park} investigate the battery management perspectives of a drone delivery business with the promise of reducing the battery and electricity costs by 25\% and the average waiting time for the packets by over 50\%.

While implementation and operation management solutions are key in enabling drone delivery, security issues and government regulations are not to be neglected. Sanjab et al.~\cite{Sanjab} introduce a mathematical framework for analyzing and enhancing the security of drone delivery systems. They approach this issue with a game theoretic analysis between the operator of a drone delivery fleet and a malicious attacker.

Finally, although public perception may be controversial regarding autonomous drone delivery, the findings of Stolaroff et al.~\cite{Stolaroff} may incite governments to pass regulations in favor of this future technology. They suggest that drone delivery has the potential to reduce energy usage and greenhouse gas emissions in the transportation sector. This claim corresponds with the vision of Google's project Wing~\cite{Wing}. Although they do not disclose implementation details, they claim that for local deliveries the CO$_2$ emissions of their aircraft are significantly lower than of delivery trucks. In contrast to our approach, the prototypes displayed on their website all include fixed wings, presumably for increased energy efficiency. Apart from designing vehicles, they are also working on an unmanned traffic management system, with the aim of enabling safe and responsible shared airspace usage for hobbyists and commercial operators alike.

Amazon's drone delivery service, Prime Air~\cite{Amazon}, tested different vehicle designs to discover the best delivery method in different operating environments. They show three different vehicle designs, but one feature is common in all of them; they include four rotors, similar to our approach.

The multinational logistics company UPS has also conducted tests with autonomous delivery drones~\cite{UPS}. However, instead of replacing delivery trucks, they plan to extend their range and increase their efficiency simultaneously with the delivery drones. The drone they tested was launched from the top of a delivery truck, autonomously delivered a package and then returned to the ground vehicle. UPS claims that reducing the path of every driver by one mile per day would result in savings of up to \$50 million.

Although each of the above mentioned multinational companies have their own approach to autonomous drone delivery, they all share one aspect; they plan to use this type of service in rural areas with low to average population density, where it is possible to navigate precisely based on GPS signals, and there is enough space for landing sites or drop off locations. In contrast, we aim to enable autonomous drone delivery in dense urban environments with buildings equipped with a small private area for landing, e.g., a balcony or a porch.

\section{Method}

In our proposed scenario, the delivery drone autonomously navigates to a location above the recipient's home using GPS based navigation. This is possible since GPS navigation remains accurate as long as the drone stays above the rooftops. Then, it switches to vision based navigation and starts descending in order to find the target balcony, tagged by a visual marker. Once the balcony is found, the drone enters the balcony, drops off the package, and leaves on the same path it arrived. Therefore, the delivery recipient needs to cooperate with the delivery service provider. On one hand, the recipient needs to print a visual marker, and stick it on the balcony door or wall where it is clearly visible. On the other hand, the recipient needs to provide GPS coordinates of a descent channel in front of the balcony, where the delivery drone can descend while scanning the building for the visual marker.

We have chosen the size of the visual marker such that it can be printed on a sheet of paper of the size US letter or A4. Our work is currently restrained to using a single marker for detection. However, in a real-life scenario the print might include two visual markers; one of which can be detected at larger distances, and another one that can encode the information necessary for parcel identification.

Once the visual marker is printed and placed on a vertical surface on the balcony, the exact GPS coordinates of a descent channel in front of the balcony need to be provided. The graphical user interface (GUI) aiding the recipients in providing the coordinates could take advantage of Google maps' high resolution satellite images in urban environments. However the implementation of the GUI is not part of this paper and left for future work. The descent channel should be free from obstacles and provide direct view on the visual marker. To further facilitate the detection of the marker, an approximate direction towards the visual marker could also be specified.

With the above mentioned prerequisites fulfilled, the delivery service provider attaches the packet to one of its autonomous drones and launches the delivery. Our algorithm assumes that the drone can navigate to the desired coordinates using GPS based navigation. Since this is only possible at high altitudes where GPS signals are not disrupted by buildings, the drone would arrive at an altitude above the delivery location.

Our algorithm uses visual-inertial localization to navigate the drone to the target balcony. First, the drone starts descending in the obstacle free channel, while continuously scanning the building for the visual marker. Once the visual marker is detected, the drone uses its trajectory planner with collision avoidance to approach the marker.

Our implementation is currently restricted to autonomously approaching the marker, however, in a final system the drone must also release the parcel and finally navigate back to the launching location on the same path it arrived.

\section{System Architecture}

\begin{figure}[t]
  \centering
  \includegraphics[width=\linewidth]{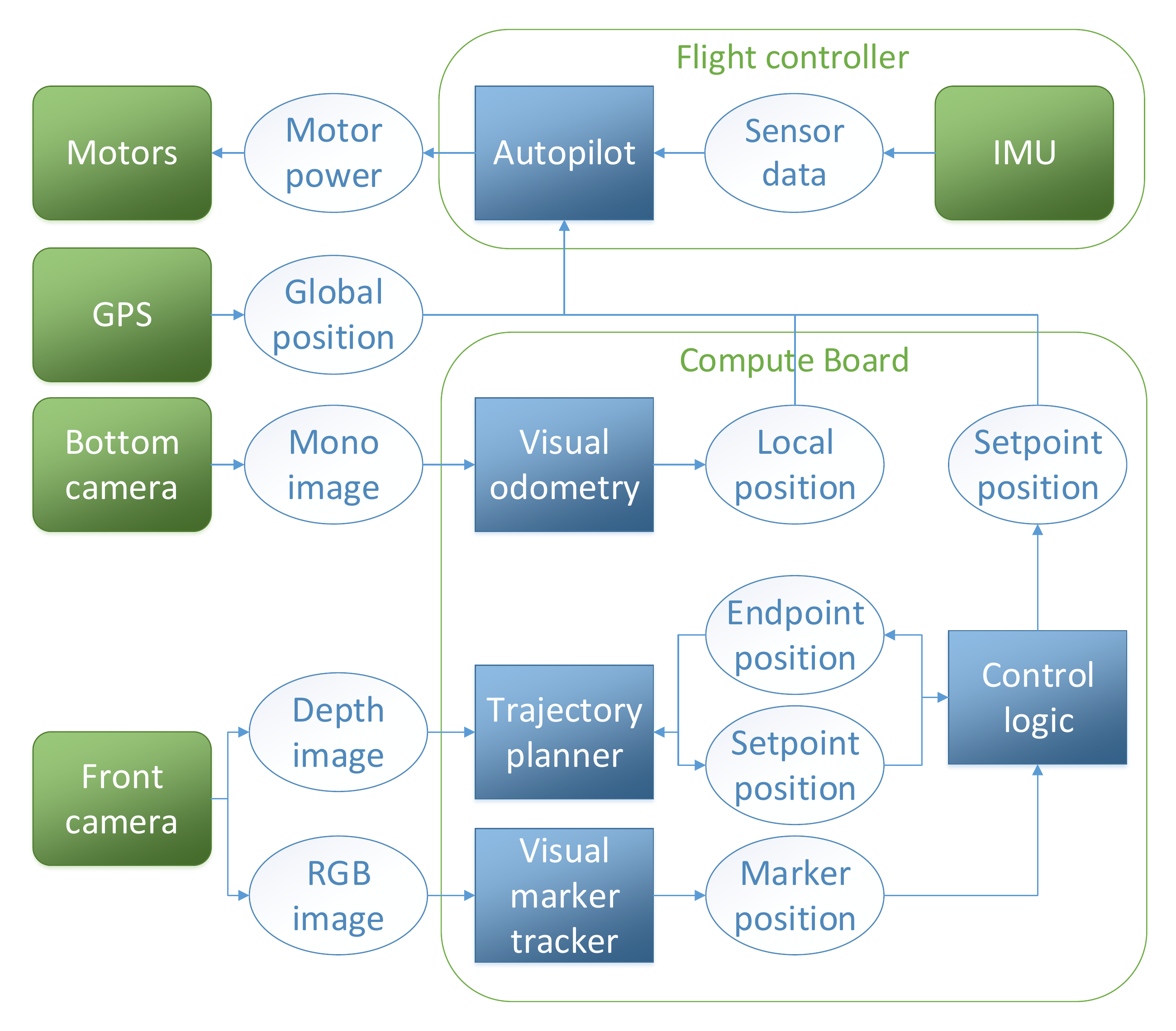}
  \caption{System architecture of the autonomous delivery drone. The hardware blocks of the system are represented by rounded green rectangles, whereas the software blocks are represented by the blue rectangles. The different modules of the system communicate with each other through the data structures described by the ellipsoids.}
  \label{fig:SysArch}
\end{figure}

Figure~\ref{fig:SysArch} shows an overview of the system architecture of our autonomous delivery drone.

The application runs on two separate processors. The flight controller contains an embedded processor capable of executing the autopilot
. All other software components of the system (i.e., visual odometry, trajectory planner, visual marker tracker and control logic) are executed on a separate processor. The autopilot can take position setpoints as its input and calculate the required motor power as the output, based on the current position estimate. The GPS module provides precise global position estimates in large open environments and thus can be used for localization with the aid of the sensor data provided by the inertial measurement unit (IMU).

However, since buildings can reflect off GPS signals causing localization errors of several meters, GPS becomes unusable for localization in clustered urban environments. Therefore, the local position provided by visual odometry is essential for this application. 
The system includes a downward facing camera providing monochrome images used for visual odometry.
It also features a front-facing camera capable of streaming depth and RGB images. The trajectory planner uses the depth images to build an occupancy grid to avoid collisions, while the visual marker tracker processes the RGB images, since we chose to identify the delivery target location with a visual marker.

Finally, the control logic block ensures autonomous mission execution. It acts as an overseer of the whole application and an interface between the two processors. Thus, it is responsible for maintaining the current state of the system (for example the marker position), executing the mission based on the system state, error handling and sending setpoint positions to the autopilot based on the mission state and trajectory planner output.

\section{Implementation}

\begin{figure}[t]
\vspace{0.2cm}
  \centering
  \includegraphics[width=\linewidth]{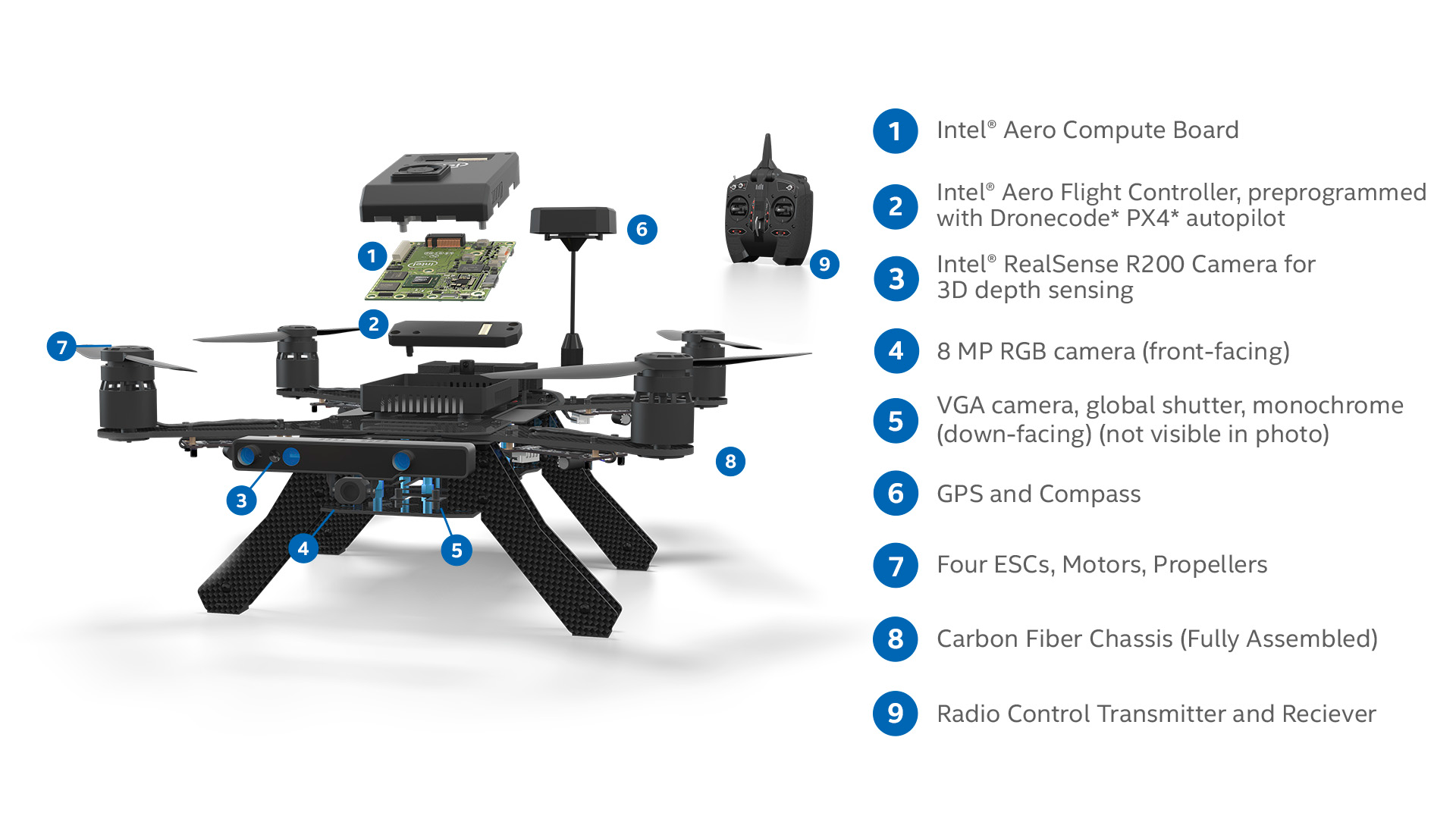}
  \caption{Hardware components of the Intel\textsuperscript{\tiny\textregistered} Aero Ready to Fly Drone~\cite{Intel}. This drone is a fully-assembled, flight tested, ready to fly quadcopter intended to alleviate some of the development burden from commercial drone developers and researchers.}
  \label{fig:RTF}
\end{figure}

Since visual odometry is a computationally intensive task, researchers in this field typically use customized drones with desktop grade processing units \cite{Surber,Papachristos,Alzugaray}. In our work we present a solution with an off-the-shelf drone, the Intel\textsuperscript{\tiny\textregistered} Aero Ready to Fly (RTF) Drone. Figure~\ref{fig:RTF} illustrates that the RTF drone contains all the hardware blocks displayed in Figure~\ref{fig:SysArch}, and is capable of executing all software blocks simultaneously.

While the autopilot runs on an embedded microprocessor (STM32) with real-time guarantees, the remaining software blocks are executed on a quad core processor with Intel x64 architecture (Intel Atom\textsuperscript{\tiny\textregistered} x7-Z8750). This processor is capable of running an Ubuntu 16.04 operating system with the Robot Operating System (ROS).

\subsection{Robot Operating System (ROS)}

Despite its name, the Robot Operating System is not an operating system but rather a middle-ware or framework to facilitate building robot applications. A ROS application consists of processes, called nodes, communicating through message queues, called topics. In order to encourage collaborative robotics software development, ROS offers a large variety of conventionalized data types to be used as topics and it is possible to define new, application specific ones.

Thanks to the open source nature of ROS, our application can take advantage of a few already existing ROS nodes developed in different laboratories all across the world. For example, we use the visual marker tracker and visual odometry ROS packages without modification, however the control logic block has been created for this application from scratch, and the trajectory planner ROS node has also been customized for our application.

Intel provides a ROS package to enable the usage of the front-facing camera of the RTF drone with ROS. Once started, various raw and processed image streams are available in ROS for further processing. There is, however, no official ROS driver for the bottom-facing camera, therefore we modified an existing ROS driver designed for V4L USB cameras in order to acquire the desired image stream.

\subsection{Autopilot}
\label{sec:Autopilot}

The flight controller of the RTF drone supports two of the most widespread autopilots; PX4~\cite{px4_paper} and Ardupilot. While both autopilots cover a great range of functionalities and are mature, well maintained software, we decided to use the PX4 flight stack for the following reasons:

\begin{itemize}

\item PX4 was meant for advanced drone applications. It can be used in hybrid systems, with safety critical algorithms running on a real-time OS that can communicate with ROS running on Linux on a companion computer~\cite{Dronesmith}.

\item It can be compiled for POSIX systems and offers a more mature software-in-the-loop (SITL) simulation with built-in support of Gazebo simulator.

\item The pose estimator in its 1.8.0 release \footnote{\url{https://github.com/PX4/Firmware/releases/tag/v1.8.0}} supports the direct fusion of data coming from visual inertial odometry and has an interface for collision avoidance.

\end{itemize}

PX4 features an off-board flight mode where the vehicle obeys a position, velocity or attitude setpoint provided through the MAVLink communication protocol. In our implementation, the compute board periodically updates the position setpoint and forwards it to the flight controller in order to achieve the autonomous navigation.

\begin{figure}[t]
  \centering
  \includegraphics[width=\linewidth]{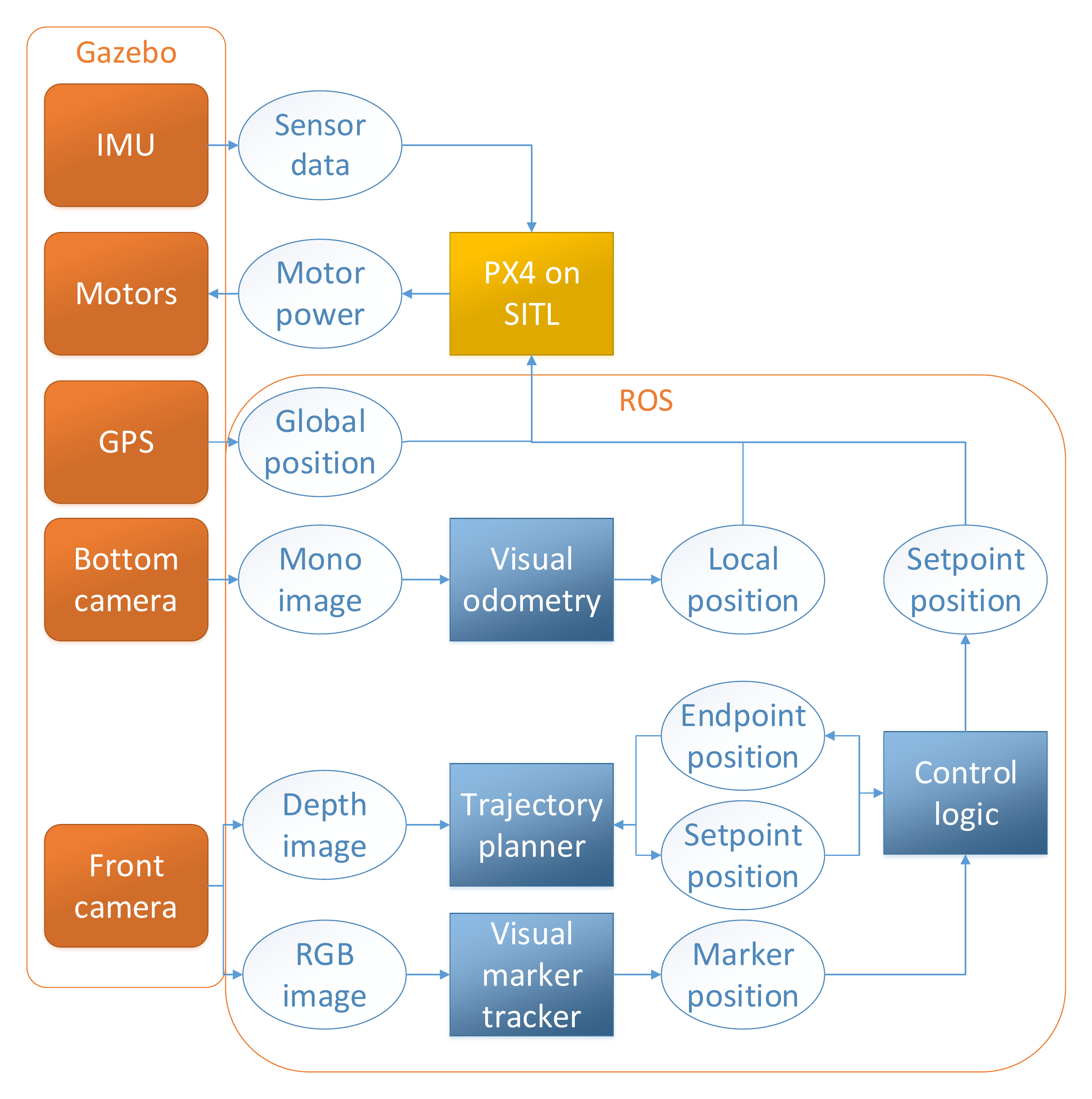}
  \caption{System architecture of the simulated autonomous delivery drone. The simulated hardware blocks of the system are represented by rounded orange rectangles, and the ROS nodes are represented by the blue rectangles. The different modules of the system communicate with each other through the data structures described by the ellipsoids.}
  \label{fig:SimArch}
\end{figure}

\subsection{Visual Odometry}
\label{sec:VIO}

We have considered several types of algorithms for vision-based localization. The options include optical flow (OF), simultaneous localization and mapping (SLAM) and visual odometry (VO).

The Intel Aero RTF drone supports the usage of OF, but requires external hardware (distance sensor and special-purpose camera) to be installed on the drone. On the other hand, a SLAM library called RTAB-Map is provided with the RTF drone installation files, but since it barely fits the power budget of the compute board, it is not usable for autonomous navigation in practice. Parallel Tracking and Mapping (PTAM)~\cite{PTAM} is a SLAM algorithm designed for hand-held cameras. While it requires low processing capacity, our empirical evaluation showed that it is not precise and robust enough to be used for vision-based localization.

The RTF drone features a monochrome downward facing camera with global shutter. This camera was meant for VO, and several mature open-source VO algorithms exist, therefore it was a logical choice to use VO for localization. Robust Visual-Inertial Odometry (ROVIO)~\cite{ROVIO} and Semi-Direct Visual Odometry (SVO)~\cite{SVO} are both well maintained open-source VO algorithms, and they both integrate with ROS. While both algorithms fit the computational budget of the compute board, the shorter processing time of SVO means shorter time delay for the pose estimator, enables faster processing rate, and thus it is more suitable for our application.

\subsection{Visual Marker Tracker}

\begin{figure}[t]
  \centering
  \includegraphics[trim=25 140 25 130, clip, width=0.4\linewidth]{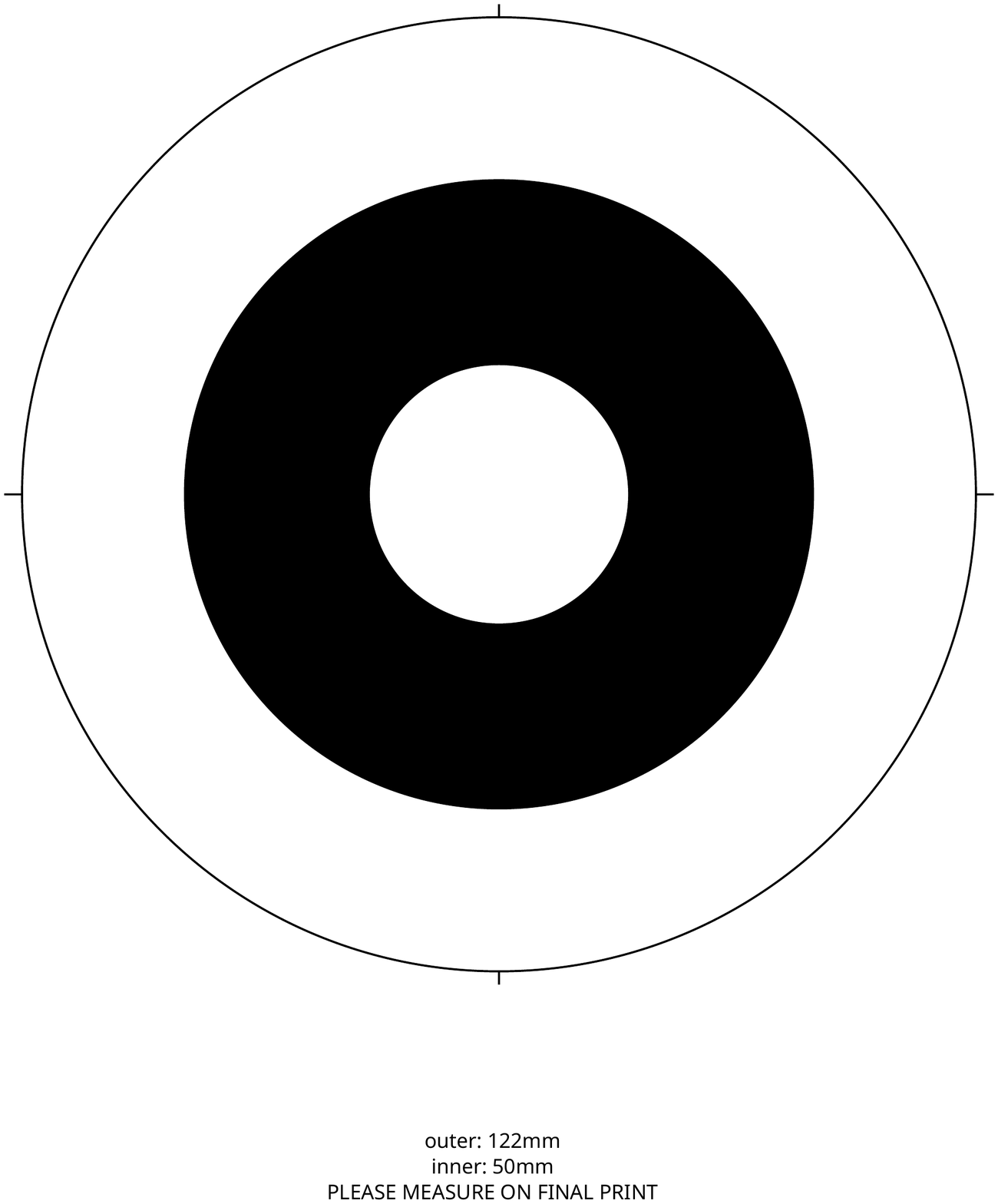}
  \caption{The WhyCon visual marker is a simple thick black circle with known inner and outer diameters. The marker print is white inside and outside around the circle, such that the black circle is clearly distinguishable from its environment.}
  \label{fig:WhyCon}
\end{figure}

Again, a large variety of visual markers with trackers exists in the literature, each with different complexities and advantages. For our application WhyCon~\cite{WhyCon} proved to be an excellent solution, since it combines advantages like precision, robustness, efficiency, low cost and it integrates with ROS. The simple design of the WhyCon visual marker, as illustrated on Figure~\ref{fig:WhyCon}, enables detection at large distances with more than one orders of magnitude smaller processing time than other marker trackers~\cite{WhyCon2}.

\subsection{Trajectory Planner} \label{sec:TrajPlan}

The main purpose of the trajectory planner is to calculate a collision free trajectory based on the current position of the drone, the desired trajectory endpoint and the image stream of the depth camera. Usenko et al.~\cite{Usenko} propose a real-time approach by storing an occupancy grid of the environment in a three-dimensional (3D) circular buffer that moves together with the drone. They represent the trajectories by uniform B-Splines, which ensures that the trajectory is sufficiently smooth and allows for efficient optimization.

Our trajectory planner node expects the coordinates of the desired target location as its input. Then, it first calculates a collision free trajectory based on the occupancy grid, with the current position of the drone as starting point and the target location as endpoint of the trajectory. Since the output of the node is a setpoint position published with a rate of 2 Hz, it also calculates uniformly distributed control points along the trajectory, subject to velocity and acceleration constraints. During flight, the occupancy grid is continuously updated based on the depth camera output. Before the next setpoint position is published towards the flight controller, the trajectory is re-optimized in order to avoid undesired behavior (e.g., collision or large deviation from the trajectory). Figure~\ref{fig:RVIZ} shows the trajectory planning in action during simulation, including the occupancy grid shown in red. 

\subsection{Control Logic}

The control logic acts as an interface between the system components and ensures autonomous mission execution. It relies on ROS topics to communicate with the remaining nodes, for example to read the estimate of the visual marker position. 
Although the autopilot uses the MAVLink communication protocol, the MAVROS package provides a communication driver and exposes various commands, system state variables and interfaces as ROS topics. For example by publishing to the corresponding MAVROS topic the autopilot can be commanded to take off and fly to a given position. Also, by implementing a callback function, the local position of the drone can be monitored and processed.
In order to keep track and conveniently switch between different coordinate frames, the control logic block uses the second generation of the transform ROS package, called \emph{tf2}.

\section{Evaluation}

Until the research community agrees on a quantitative yet meaningful evaluation method for drone delivery applications, empirical evaluation remains the best option. Since field tests are expensive both in terms of time and - in case of crashes - money, a simulation environment is essential for development and testing.

\subsection{Simulation}

\begin{figure}[t]
\vspace{0.2cm}
  \centering
  \includegraphics[trim=0 100 0 150, clip, width=\linewidth]{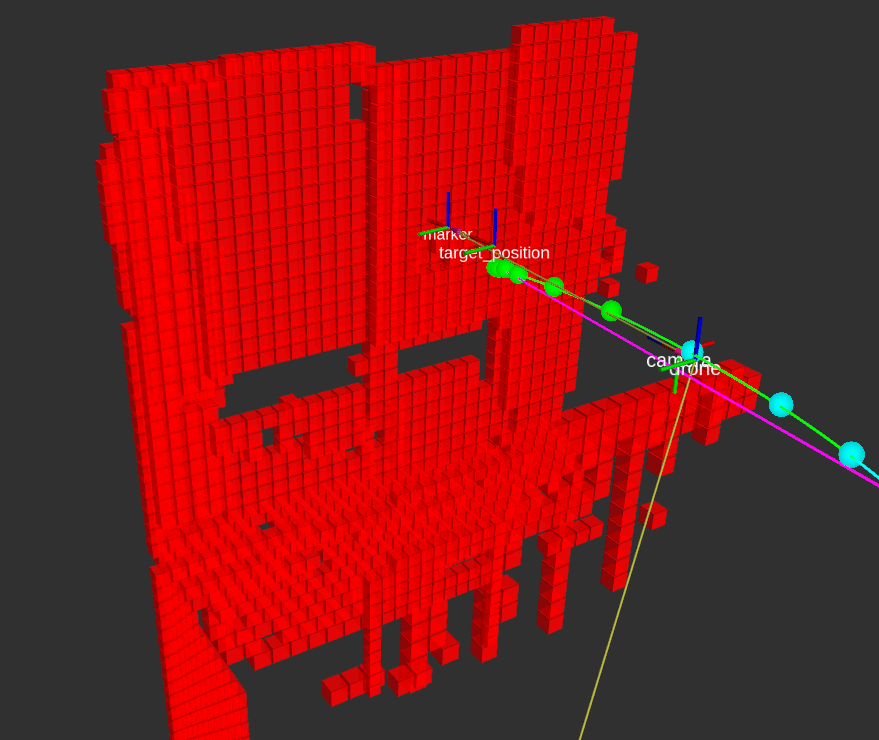}
\caption{RVIZ GUI snapshot of the same environment as seen in Figure~\ref{fig:Gazebo}. The red cubes represent the occupancy grid, while the blue and green dots along the green trajectory line are the past and future control setpoints sent to the flight controller. The perpendicular red-green-blue lines with texts represent different frames logged by the transform package.}
  \label{fig:RVIZ}
\end{figure}

As mentioned in Section~\ref{sec:Autopilot}, PX4 offers SITL simulation with support for the Gazebo simulator. Thus, it is possible to simulate the application on a single computer running Ubuntu.
Figure~\ref{fig:SimArch} illustrates the system architecture of the simulated autonomous delivery drone. Compared to Figure~\ref{fig:SysArch} it is apparent that the ROS software nodes are left untouched. Thus, they can be launched and executed on the computer used for simulation in the same manner as they were on the compute board of the drone. The main difference is that they receive their input image streams from simulated cameras inside the Gazebo simulator, instead of the real hardware. The same applies for the autopilot; PX4 on SITL communicates with ROS through MAVROS messages the same way it would if it were executed inside the flight controller.

\subsection{Visualization}\label{subsec:visualization}

\begin{figure}[t]
\vspace{0.2cm}
  \centering
  \includegraphics[trim=200 100 100 100, clip, width=\linewidth]{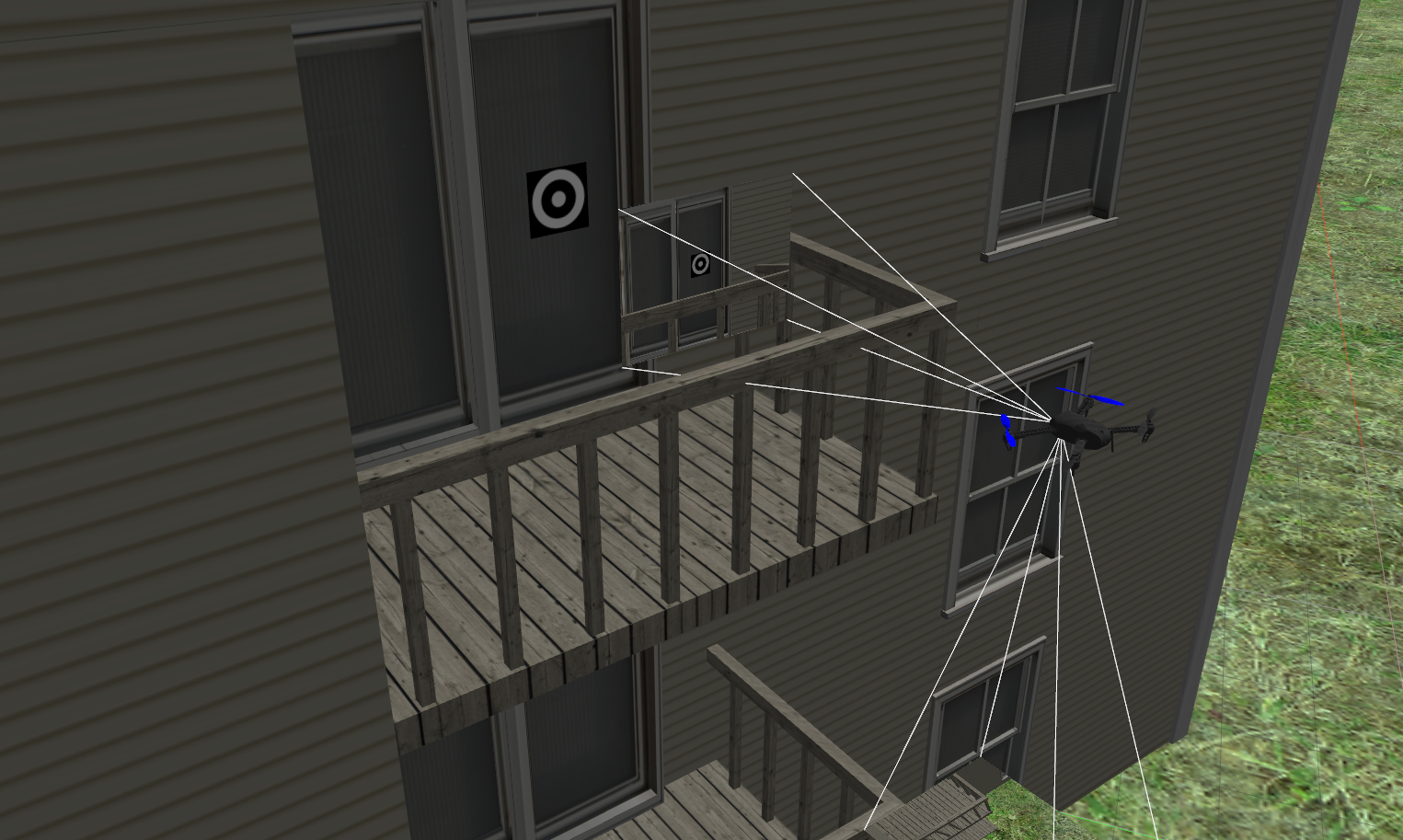}
  \caption{Snapshot of the Gazebo simulator GUI during autonomous delivery mission execution. The simulated drone has currently found the visual marker and attempts to approach it using the trajectory planner with collision avoidance.}
  \label{fig:Gazebo}
\end{figure}

Gazebo offers a GUI for 3D visualization of the simulation environment. RVIZ on the other hand is a tool provided by ROS offering 3D visualization of transformations and sensor information, as illustrated in Figure~\ref{fig:RVIZ}. While the GUI of Gazebo displays how a human perceives the environment, RVIZ visualizes the robot's sensory information, and can be used with real sensor data as well.
With the help of RVIZ we visualize the occupancy grid stored in the the ring buffer, as described in Section~\ref{sec:TrajPlan}. We also visualize the current trajectory of the drone, including the past and future control setpoints sent to the flight controller. Finally, we visualize various frames and positions as maintained by the transform package, like the drone body frame and the estimated position of the visual marker.

\subsection{Experimental Results}

We evaluate the maximum detection distance and processing time of the WhyCon visual marker tracker with both image resolutions supported by the front-facing camera of the drone. For the distance measurements the marker was moved along the optical axis of the camera, however similar distances were measured with the marker on the edge of the camera field of view. At larger detection distances the visual marker could be tilted up to 45 degrees without detection failure.
The processing times include the overhead introduced by the message propagation of ROS and represent the total time elapsed between capturing an image and receiving the position of the visual marker. The results in Table~\ref{tbl:WhyCon} show that although increasing the image resolution can double the maximal detection distance, the average processing time increases nearly tenfold.

\begin{figure}[t]
  \centering
  \includegraphics[width=0.6\linewidth]{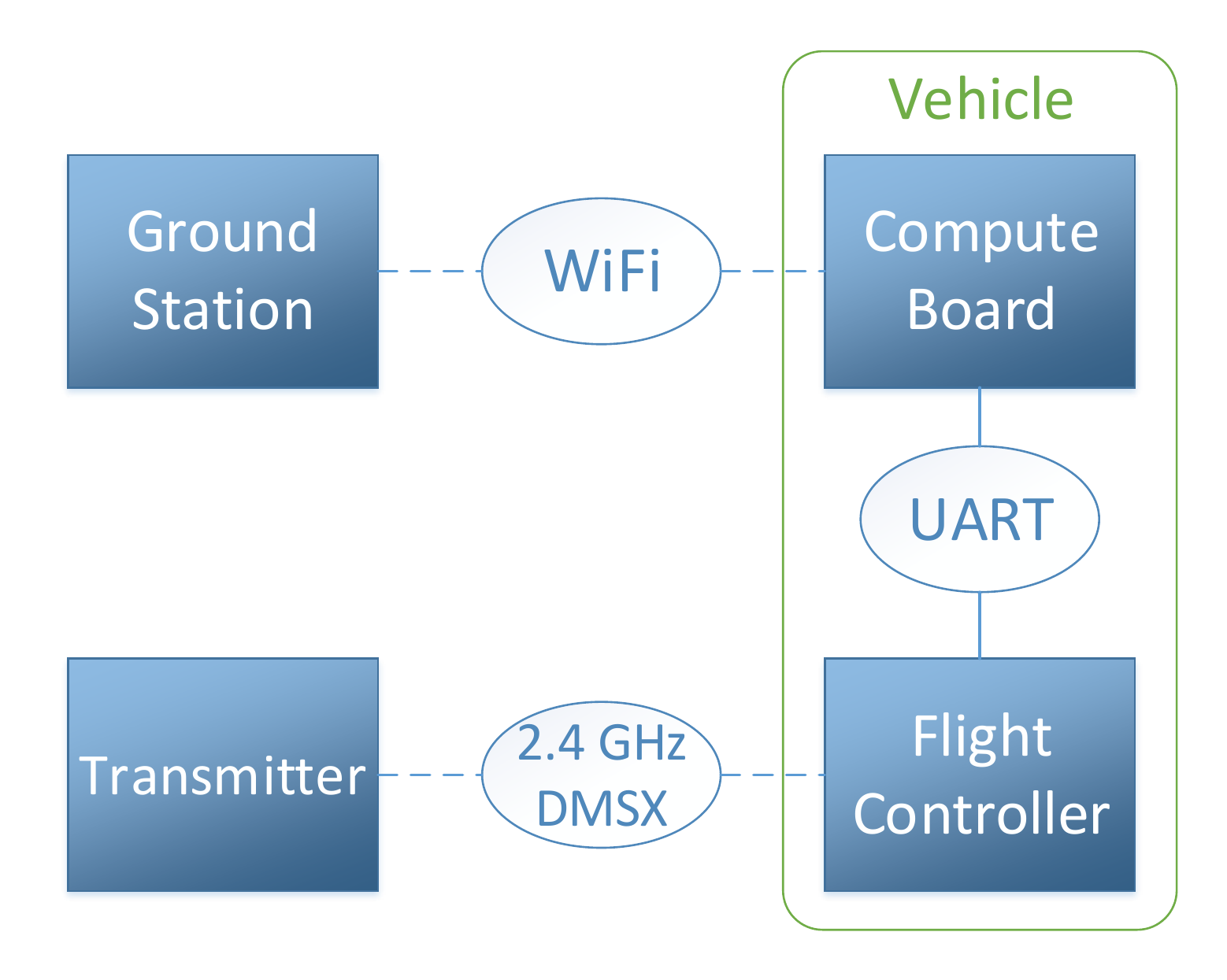}
  \caption{Overview of the field test setup. The vehicle can be controlled in manual flight mode using the transmitter, or in off-board flight mode through MAVLink commands sent by the compute board. Code execution on the compute board can be launched and monitored on a ground station through WiFi connection.}
  \label{fig:TestSetup}
\end{figure}

\begin{table}[b]
	\caption{Evaluation of WhyCon visual marker tracker on the Intel Aero Compute Board}
	\label{tbl:WhyCon}
    \begin{center}
	\begin{tabular}{|c|c|cc|}
		\hline
		Image size & Maximal detection & \multicolumn{2}{|c|}{Processing time [ms]} \\
		{[pixels x pixels]} & distance [m] & Avg & Max \\
		\hline
		640 x 480 & 7 $\pm$ 0.5 & 10.61 & 12.23 \\
		1920 x 1080 & 14 $\pm$ 0.5 & 102.35 & 240.72 \\
		\hline
	\end{tabular}
    \end{center}
\end{table}

We also evaluated the processor usage of three different vision-based localization algorithms on the compute board. The processor usage of the whole system was measured with the Unix program \emph{htop} over a one minute time period, while executing different vision-based localization algorithms. The results are summarized by Table~\ref{tbl:VIO}. All algorithms were executed with an input image stream of 640 by 480 monochrome pixels published at a rate of 30 Hz. The results show that PTAM has the lowest average processor usage per core. Although the processor usage of ROVIO is significantly lower than of SVO, SVO makes better use of the 4 available processor cores and processes the frames significantly faster. Even after reducing the performance settings of ROVIO (NMAXFEATURE=15, PATCHSIZE=4, NPOSE=0), the long processing time per frame resulted in an average loss of 5 frames per second. Therefore, in the final implementation we use SVO.

\begin{table}[b]
	\caption{Average processor utilization of the compute board over a 1 minute period while executing different vision-based localization algorithms}
	\label{tbl:VIO}
    \begin{center}
	\begin{tabular}{|c|c|c|}
		\hline
		VIO algorithm & Avg. processor usage [\%] & Processing rate [Hz] \\
		\hline
        Without & 8 $\pm$ 5 & n/a \\
		PTAM & 22 $\pm$ 10 & 30 \\
		SVO & 71 $\pm$ 10 & 30 \\
		ROVIO (default) & 51 $\pm$ 10 & 15 \\
		ROVIO (reduced) & 46 $\pm$ 10 & 25 \\
		\hline
	\end{tabular}
    \end{center}
\end{table}

\subsection{Field Test}

\begin{figure}[t]
\vspace{0.2cm}
  \centering
  \includegraphics[width=0.9\linewidth]{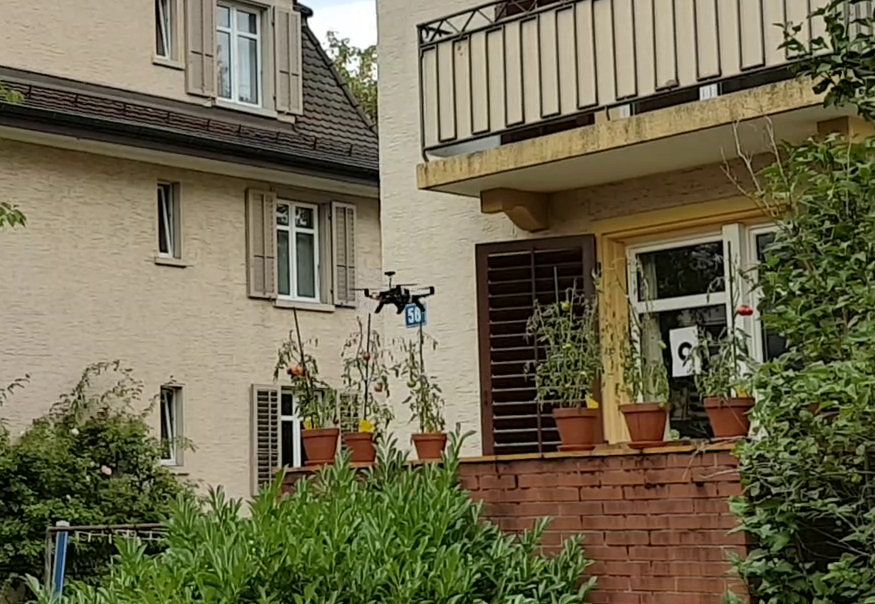}
  \caption{Snapshot of the drone approaching the visual marker on a balcony during a field test.}
  \label{fig:FieldTest}
\end{figure}

Figure~\ref{fig:TestSetup} displays the setup needed for testing our autonomous delivery drone application. The compute board of the vehicle can act as a WiFi hotspot and thus enables a laptop, or any WiFi compatible device to establish a connection with it. Then, the ground station can connect to the compute board through ssh and launch the ROS nodes of the application. The ground station can also run the \emph{qGroundControl} application, which communicates with the flight controller (through the compute board) and provides numerous features like full flight control, autopilot parameter changes and flight log recordings. Figure~\ref{fig:FieldTest} shows our drone while approaching the visual marker on an urban balcony. A link to a complete video demonstrating the functionality of our code can be found on our GitHub repository referenced in Section~\ref{sec:Intro}.
Note that it is recommended to bind one of the switch positions on the radio transmitter to manual flight mode. In case the developer observes unexpected behavior she can immediately regain control over the drone by switching to manual flight mode.

\section{Conclusion and Future Work}

Although autonomous drone delivery might sound like science fiction today, our work provides a working proof of concept that can be further extended to a complete last mile delivery solution in urban environments. Despite the complexity of this problem, we were able to address several key aspects of it including vision-based navigation and target location detection.

Additional fundamental issues include designing a custom vehicle that suits our application while being energy efficient, quiet and safe. It is also not trivial how the delivery drone should transport and release the parcel upon arrival (e.g., simply drop it, lower it on a rope). One would also need to implement security measurements against malicious attacks, such as jamming or spoofing. Finally the solution needs to be scaled to a delivery drone fleet (operation management).

The repository containing our algorithms and calibration data is publicly available on GitHub including detailed installation and usage instructions. We believe this repository can form the basis of various research projects investigating the above mentioned topics or exploring the possibilities of autonomous drone applications in general.








\cleardoublepage
\newpage
\nocite*{} 
\bibliography{IEEEabrv,./bibliography}

\begin{thebibliography}{10}
\providecommand{\url}[1]{#1}
\csname url@samestyle\endcsname
\providecommand{\newblock}{\relax}
\providecommand{\bibinfo}[2]{#2}
\providecommand{\BIBentrySTDinterwordspacing}{\spaceskip=0pt\relax}
\providecommand{\BIBentryALTinterwordstretchfactor}{4}
\providecommand{\BIBentryALTinterwordspacing}{\spaceskip=\fontdimen2\font plus
\BIBentryALTinterwordstretchfactor\fontdimen3\font minus
  \fontdimen4\font\relax}
\providecommand{\BIBforeignlanguage}[2]{{%
\expandafter\ifx\csname l@#1\endcsname\relax
\typeout{** WARNING: IEEEtran.bst: No hyphenation pattern has been}%
\typeout{** loaded for the language `#1'. Using the pattern for}%
\typeout{** the default language instead.}%
\else
\language=\csname l@#1\endcsname
\fi
#2}}
\providecommand{\BIBdecl}{\relax}
\BIBdecl

\bibitem{worldbankurbanpop}
W.~Bank, ``Ups tests residential delivery via drone launched from atop package
  car,'' \url{https://data.worldbank.org/indicator/SP.URB.TOTL.IN.ZS}, 2018,
  accessed: 2018-09-15.

\bibitem{Joerss}
M.~Joerss, A.~Schröder, F.~Neuhaus, C.~Klink, and F.~T. Mann, ``Parcel
  delivery: The future of the last mile,'' McKinsey \& Company, Tech. Rep.,
  2016.

\bibitem{Andrea}
\BIBentryALTinterwordspacing
R.~D'Andrea, ``Guest editorial can drones deliver?'' \emph{{IEEE} Trans.
  Automation Science and Engineering}, vol.~11, no.~3, pp. 647--648, 2014.
  [Online]. Available: \url{https://doi.org/10.1109/TASE.2014.2326952}
\BIBentrySTDinterwordspacing

\bibitem{Guerrero}
\BIBentryALTinterwordspacing
M.~E. Guerrero, D.~A. Mercado, R.~Lozano, and C.~D. Garcia, ``Passivity based
  control for a quadrotor {UAV} transporting a cable-suspended payload with
  minimum swing,'' in \emph{54th {IEEE} Conference on Decision and Control,
  {CDC} 2015, Osaka, Japan, December 15-18, 2015}, 2015, pp. 6718--6723.
  [Online]. Available: \url{https://doi.org/10.1109/CDC.2015.7403277}
\BIBentrySTDinterwordspacing

\bibitem{Dorling}
\BIBentryALTinterwordspacing
K.~Dorling, J.~Heinrichs, G.~G. Messier, and S.~Magierowski, ``Vehicle routing
  problems for drone delivery,'' \emph{{IEEE} Trans. Systems, Man, and
  Cybernetics: Systems}, vol.~47, no.~1, pp. 70--85, 2017. [Online]. Available:
  \url{https://doi.org/10.1109/TSMC.2016.2582745}
\BIBentrySTDinterwordspacing

\bibitem{Gatteschi}
\BIBentryALTinterwordspacing
V.~Gatteschi, F.~Lamberti, G.~Paravati, A.~Sanna, C.~Demartini, A.~Lisanti, and
  G.~Venezia, ``New frontiers of delivery services using drones: {A} prototype
  system exploiting a quadcopter for autonomous drug shipments,'' in \emph{39th
  {IEEE} Annual Computer Software and Applications Conference, {COMPSAC} 2015,
  Taichung, Taiwan, July 1-5, 2015. Volume 2}, 2015, pp. 920--927. [Online].
  Available: \url{https://doi.org/10.1109/COMPSAC.2015.52}
\BIBentrySTDinterwordspacing

\bibitem{Park}
\BIBentryALTinterwordspacing
S.~Park, L.~Zhang, and S.~Chakraborty, ``Battery assignment and scheduling for
  drone delivery businesses,'' in \emph{2017 {IEEE/ACM} International Symposium
  on Low Power Electronics and Design, {ISLPED} 2017, Taipei, Taiwan, July
  24-26, 2017}, 2017, pp. 1--6. [Online]. Available:
  \url{https://doi.org/10.1109/ISLPED.2017.8009165}
\BIBentrySTDinterwordspacing

\bibitem{Sanjab}
\BIBentryALTinterwordspacing
A.~Sanjab, W.~Saad, and T.~Basar, ``Prospect theory for enhanced cyber-physical
  security of drone delivery systems: {A} network interdiction game,'' in
  \emph{{IEEE} International Conference on Communications, {ICC} 2017, Paris,
  France, May 21-25, 2017}, 2017, pp. 1--6. [Online]. Available:
  \url{https://doi.org/10.1109/ICC.2017.7996862}
\BIBentrySTDinterwordspacing

\bibitem{Stolaroff}
J.~Stolaroff, C.~Samaras, E.~O’Neill, A.~Lubers, A.~Mitchell, and
  D.~Ceperley, ``Energy use and life cycle greenhouse gas emissions of drones
  for commercial package delivery,'' \emph{Nature Communications}, vol.~9,
  no.~1, 2018.

\bibitem{Wing}
X, ``Our approach – wing,'' \url{https://x.company/wing/our-approach}, 2018,
  accessed: 2018-09-15.

\bibitem{Amazon}
Amazon, ``Amazon prime air,''
  \url{https://www.amazon.com/Amazon-Prime-Air/b?ie=UTF8\&node=8037720011},
  2018, accessed: 2018-09-15.

\bibitem{UPS}
UPS, ``Ups tests residential delivery via drone launched from atop package
  car,''
  \url{https://pressroom.ups.com/pressroom/ContentDetailsViewer.page?ConceptType=PressReleases\&id=1487687844847-162},
  2018, accessed: 2018-09-15.

\bibitem{Surber}
\BIBentryALTinterwordspacing
J.~Surber, L.~Teixeira, and M.~Chli, ``Robust visual-inertial localization with
  weak {GPS} priors for repetitive {UAV} flights,'' in \emph{2017 {IEEE}
  International Conference on Robotics and Automation, {ICRA} 2017, Singapore,
  Singapore, May 29 - June 3, 2017}, 2017, pp. 6300--6306. [Online]. Available:
  \url{https://doi.org/10.1109/ICRA.2017.7989745}
\BIBentrySTDinterwordspacing

\bibitem{Papachristos}
C.~Papachristos, S.~Khattak, and K.~Alexis, ``Autonomous exploration of
  visually-degraded environments using aerial robots,'' in \emph{2017
  International Conference on Unmanned Aircraft Systems (ICUAS)}, June 2017,
  pp. 775--780.

\bibitem{Alzugaray}
\BIBentryALTinterwordspacing
I.~Alzugaray, L.~Teixeira, and M.~Chli, ``Short-term {UAV} path-planning with
  monocular-inertial {SLAM} in the loop,'' in \emph{2017 {IEEE} International
  Conference on Robotics and Automation, {ICRA} 2017, Singapore, Singapore, May
  29 - June 3, 2017}, 2017, pp. 2739--2746. [Online]. Available:
  \url{https://doi.org/10.1109/ICRA.2017.7989319}
\BIBentrySTDinterwordspacing

\bibitem{Intel}
Intel, ``Intel aero ready to fly drone,''
  \url{https://www.intel.com/content/www/us/en/products/drones/aero-ready-to-fly.html},
  2018, accessed: 2018-09-15.

\bibitem{px4_paper}
\BIBentryALTinterwordspacing
L.~Meier, D.~Honegger, and M.~Pollefeys, ``{PX4:} {A} node-based multithreaded
  open source robotics framework for deeply embedded platforms,'' in
  \emph{{IEEE} International Conference on Robotics and Automation, {ICRA}
  2015, Seattle, WA, USA, 26-30 May, 2015}, 2015, pp. 6235--6240. [Online].
  Available: \url{https://doi.org/10.1109/ICRA.2015.7140074}
\BIBentrySTDinterwordspacing

\bibitem{Dronesmith}
\BIBentryALTinterwordspacing
D.~Technologies, ``Why we chose px4 (vs apm) as luci’s default firmware.''
  [Online]. Available:
  \url{https://medium.com/@Dronesmith/why-we-chose-px4-vs-apm-as-lucis-default-firmware-ea39f4514bef}
\BIBentrySTDinterwordspacing

\bibitem{PTAM}
\BIBentryALTinterwordspacing
G.~Klein and D.~W. Murray, ``Parallel tracking and mapping for small {AR}
  workspaces,'' in \emph{Sixth {IEEE/ACM} International Symposium on Mixed and
  Augmented Reality, {ISMAR} 2007, 13-16 November 2007, Nara, Japan}, 2007, pp.
  225--234. [Online]. Available:
  \url{https://doi.org/10.1109/ISMAR.2007.4538852}
\BIBentrySTDinterwordspacing

\bibitem{ROVIO}
\BIBentryALTinterwordspacing
M.~Bloesch, M.~Burri, S.~Omari, M.~Hutter, and R.~Siegwart, ``Iterated extended
  kalman filter based visual-inertial odometry using direct photometric
  feedback,'' \emph{I. J. Robotics Res.}, vol.~36, no.~10, pp. 1053--1072,
  2017. [Online]. Available: \url{https://doi.org/10.1177/0278364917728574}
\BIBentrySTDinterwordspacing

\bibitem{SVO}
\BIBentryALTinterwordspacing
C.~Forster, Z.~Zhang, M.~Gassner, M.~Werlberger, and D.~Scaramuzza, ``{SVO:}
  semidirect visual odometry for monocular and multicamera systems,''
  \emph{{IEEE} Trans. Robotics}, vol.~33, no.~2, pp. 249--265, 2017. [Online].
  Available: \url{https://doi.org/10.1109/TRO.2016.2623335}
\BIBentrySTDinterwordspacing

\bibitem{WhyCon}
M.~Nitsche, T.~Krajn{\'\i}k, P.~{\v C}{\' i}{\v z}ek, M.~Mejail, and
  T.~Duckett, ``Whycon: An efficent, marker-based localization system,'' 2015.

\bibitem{WhyCon2}
\BIBentryALTinterwordspacing
T.~Krajn{\'{\i}}k, M.~A. Nitsche, J.~Faigl, P.~Vanek, M.~Saska, L.~Preucil,
  T.~Duckett, and M.~Mejail, ``A practical multirobot localization system,''
  \emph{Journal of Intelligent and Robotic Systems}, vol.~76, no. 3-4, pp.
  539--562, 2014. [Online]. Available:
  \url{https://doi.org/10.1007/s10846-014-0041-x}
\BIBentrySTDinterwordspacing

\bibitem{Usenko}
\BIBentryALTinterwordspacing
V.~C. Usenko, L.~von Stumberg, A.~Pangercic, and D.~Cremers, ``Real-time
  trajectory replanning for mavs using uniform b-splines and a 3d circular
  buffer,'' in \emph{2017 {IEEE/RSJ} International Conference on Intelligent
  Robots and Systems, {IROS} 2017, Vancouver, BC, Canada, September 24-28,
  2017}, 2017, pp. 215--222. [Online]. Available:
  \url{https://doi.org/10.1109/IROS.2017.8202160}
\BIBentrySTDinterwordspacing

\bibitem{PX4}
PX4, ``Offboard - px4 user guide,''
  \url{https://docs.px4.io/en/flight\_modes/offboard.html}, 2018, accessed:
  2018-09-15.

\end{thebibliography}
\bibliographystyle{IEEEtran}

\end{document}